\documentclass[AMS,STIX2COL]{WileyNJDv5} 

\usepackage{pdflscape} 
\usepackage{colortbl}  
\usepackage{tikz}      
\usepackage{pdfpages}  

\usepackage[absolute,overlay]{textpos}
\usepackage{xcolor}

\newcommand{\preprintnotice}{
\begin{textblock*}{4.5cm}(1.25cm,14cm) 
\footnotesize
\setlength{\fboxsep}{4pt}
\color{black}
\fbox{
\parbox{4.5cm}{
\textbf{Preprint Notice}\
This manuscript is currently under review at
\emph{Healthcare Technology Letters} (Wiley/IET).
This version has not undergone peer review.
}
}
\end{textblock*}
}

\articletype{Letter}%

\received{Date Month Year}
\revised{Date Month Year}
\accepted{Date Month Year}
\journal{Journal}
\volume{00}
\copyyear{2023}
\startpage{1}

\begin{document}
\preprintnotice
\title{Expert Consensus on Criteria for the Automated Assessment of Laparoscopic Camera Navigation}

\author[1]{Amir Ebrahimzadeh}

\author[1]{Nazila Esmaeili}

\author[1]{Michael Ghadimi}

\author[1,2]{Jannis Hagenah}

\authormark{EBRAHIMZADEH \textsc{et al.}}
\titlemark{EXPERT CONSENSUS ON CRITERIA FOR THE AUTOMATED ASSESSMENT OF LAPAROSCOPIC CAMERA NAVIGATION}

\address[1]{\orgdiv{Center for Digital Surgery, Department of General, Visceral and Pediatric Surgery}, \orgname{University Medical Center Göttingen}, \orgaddress{\street{Robert-Koch-Straße 40}, \city{Göttingen}, \postcode{37075}, \country{Germany}}}

\address[2]{\orgname{Fraunhofer Research Institution for Individualized and Cell-based Medical Engineering (IMTE)}, \orgaddress{\street{Mönkhofer Weg 239a}}, \city{Lübeck}, \postcode{23562}, \country{Germany}}

\corres{Amir Ebrahimzadeh, \email{amir.ebrahimzadeh@med.uni-goettingen.de}}

\presentaddress{\orgdiv{Center for Digital Surgery, Department of General, Visceral and Pediatric Surgery}, \orgname{University Medical Center Göttingen}, \orgaddress{\street{Robert-Koch-Straße 40}, \city{Göttingen}, \postcode{37075}, \country{Germany}}}

\abstract[Abstract]{
\textbf{Background:} Laparoscopic camera navigation (LCN) is a critical skill, yet its current assessment typically relies on manual rating systems which are time-consuming and difficult to scale. Automated feedback could significantly enhance surgical training by providing immediate, standardized metrics. This study aims to define, clinically evaluate the relevance, and establish the technical readiness of a set of approaches for LCN assessment.

\textbf{Methods:} We developed a detailed taxonomy of 14 key aspects of camera navigation, categorized into \emph{Framing~\&~Composition}, \emph{Visibility~\&~Clarity}, \emph{Orientation~\&~Stability}, \emph{Motion~\&~Dynamics}, and \emph{Safety~\&~Awareness}. For each aspect, we assessed the technological readiness of automated measurement based on the current state of the art (SoTA) in computer vision (CV). To establish clinical relevance, we designed a survey for practicing laparoscopic surgeons to rate the importance of each aspect on a 5-point Likert scale and to select the five most critical skills.

\textbf{Results:} 23 surgeons participated in the survey. Foundational aspects like \emph{Field of View}, \emph{Focus} and \emph{Centering} were rated as most important by surgeons. We present a "Clinical Importance vs. CV Technological Readiness" matrix, identifying high-priority targets for development--aspects that are both clinically crucial and technologically ready to measure.

\textbf{Conclusion:} This work establishes a foundational framework for quantifying LCN skills. By aligning surgeon priorities with CV capabilities, we provide a clear roadmap for automatic skill assessment. This foundation enables the development of AI-driven assistance tools that can accelerate the learning curve for surgical assistants and potentially improve surgical safety and efficiency.
}


\maketitle

\section{Introduction}

Minimally invasive procedures, such as laparoscopic surgery, rely entirely on the laparoscopic camera to mediate the surgeon's visual perception and motor actions. Consequently, visual field quality and stability are critical determinants of operative safety, efficiency, and ergonomics\cite{Huettl.2020}. Effective laparoscopic camera navigation (LCN) provides a consistently clear, centered, and well-oriented view, facilitating precise manipulation. Conversely, suboptimal LCN can obscure anatomy, prolong operative time, and elevate intraoperative complication risks \cite{Nilsson.2017,Huettl.2020}. Poor navigation contributes significantly to patient injury; up to 72\% of visualization-related reports link to patient harm \cite{Ameerah.2025}. Furthermore, surgeons spend an estimated 40\% of operating time under suboptimal visual conditions, contributing to nearly 20\% of surgical complications \cite{Dhingra.2025,Lacki.2025}. While clinically important, LCN assessment remains challenging due to a lack of scalable, automated evaluation methods.

LCN utilizes several modalities, each presenting distinct advantages and limitations. Traditionally, the camera is manually controlled by a human assistant--often a trainee--whose adaptability is beneficial but subject to fatigue, miscommunication, and considerable performance variability \cite{Huber.2018}. Technological alternatives seek to address these issues.

Robotic camera holders offer stable, tremor-free images but may impose additional cognitive load, as the surgeon must manually adjust the view via hand or foot controls \cite{Wan.2023}.

In fully integrated robotic-assisted systems, such as the da Vinci Surgical System, the surgeon controls both instruments and the camera, eliminating communication latency but concentrating responsibilities on a single operator. Moreover, high financial and logistical demands restrict routine clinical accessibility.

Across these modalities--human-assisted, robotic, or hybrid (e.g., human-guided robotic holders)--fundamental principles of effective navigation remain constant. However, LCN evaluation typically relies on manual expert observation. This inherently time-consuming approach demands experienced surgeons, whose availability is limited. Dependence on manual scoring hinders widespread implementation of standardized feedback and limits potential for data-driven curriculum design.

Structured assessment tools, such as the Structured Assessment of Laparoscopic Assistant Skills (SALAS) score, provide a foundation by quantifying basic attributes like image centering and horizon alignment. However, these tools rely on manual observer ratings, restricting scalability \cite{Huber.2018}. Similarly, the Objective Structured Assessment of Camera Navigation Skills (OSA-CNS) utilizes a human-rated 5-point scale \cite{Nilsson.2017}. A critical limitation is their design for human interpretation, overlooking high-frequency kinematic and motion-based features quantifiable by modern computer vision (CV) techniques. Enabling automated skill assessment requires a framework defining LCN categories that are both clinically relevant and computationally solvable.

To address these gaps, this study proposes a consensus-driven, clinically grounded framework for LCN assessment. We pursue three objectives: 1. Conceptual decomposition: Deconstruct "good" LCN into a taxonomy of distinct, definable components. 2. Clinical prioritization: Establish relative importance through expert consensus among laparoscopic surgeons. 3. Technical readiness: Evaluate readiness of quantifying these components using SoTA CV literature.

\subsection{Contribution of this work}

This work bridges the gap between clinical requirements and technical capabilities. We provide a roadmap for developing automated assessment tools by identifying LCN skills that are high-priority targets--those deemed essential by surgeons and ready for current CV algorithms. This prioritization serves as a foundation for standardized evaluation, training, and future development of intelligent surgical assistance systems.

\section{Materials and Methods}\label{s2.methods}

Objective assessment and automated assistance for LCN are impeded by the absence of a standardized definition of requisite skills. To address this gap, this study establishes a foundational framework through a three-phase methodology: (i) proposing a taxonomy of LCN skills; (ii) evaluating the clinical relevance of this taxonomy via a survey of laparoscopic surgeons; and (iii) evaluating the technical readiness of capable CV-based quantification approaches based on SoTA capabilities.

\subsection{A Multi-faceted Taxonomy of LCN Skills}

To provide a structured basis for LCN assessment, we established a skill taxonomy comprising 14 measurable aspects. These aspects aim to capture essential components of effective camera work and, to the extent possible, are designed to be mutually exclusive. For clarity, these 14 skills are classified into five primary super-categories:
\begin{itemize}
\item \textbf{Framing \& Composition:} Pertains to geometric content, ensuring relevant instruments and anatomical structures are appropriately positioned, centered, and magnified within the frame.
\item \textbf{Visibility \& Clarity:} Concerns optical quality, focusing on maintaining sharp focus, adequate illumination, and an unobstructed field free of artifacts such as smoke or lens contamination.
\item \textbf{Orientation \& Stability:} Addresses maintaining a stable and ergonomically sound viewpoint, critical for surgeons' spatial orientation and minimizing visual fatigue.
\item \textbf{Motion \& Dynamics:} Evaluates the quality of camera movements, emphasizing smoothness, efficiency, and responsiveness to the primary surgeon's actions, including proactive repositioning.
\item \textbf{Safety \& Awareness:} Encompasses higher-level skills reflecting procedural understanding, such as avoiding collisions and maintaining a view appropriate for the broader surgical context.
\end{itemize}

We developed the LCN taxonomy by identifying and extending the core domains of established manual observer scales in the literature, specifically SALAS and OSA-CNS \cite{Huber.2018,Nilsson.2017}. To adapt and expand these domains for automated computer vision applications, we introduced additional metrics designed to capture continuous variables. This expansion incorporated (i) optical quality metrics, such as Focus, Lighting \& Exposure, and Lens Cleanliness, and (ii) high-frequency motion kinematics, such as Smoothness and Economy of Motion. While these features are challenging for human observers to evaluate in real time, they are quantifiable using CV pipelines.

To minimize conceptual redundancy, we aimed to define each aspect as mutually exclusive as possible. In practice, however, these aspects are deeply intertwined. For example, maintaining an adequate Field of View is a necessary but insufficient condition for proper Centering, as the area of surgical action must first be visible within the frame before it can be centered. Similarly, survey participants might experience difficulty distinguishing other closely related concepts, such as separating Magnification from Focus. While we provided precise definitions in the survey to mitigate these ambiguities, some conceptual overlap remains an inherent limitation of this multi-faceted taxonomy.

A detailed description of each specific skill is provided in Table~\ref{t.overview}.

\begin{table*}[htbp]
\centering
\renewcommand{\multirowsetup}{\centering}
\renewcommand{\arraystretch}{1.3}
\caption{Overview over all aspects that constitute LCN.}
\label{t.overview}
\footnotesize
\begin{tabular}{ p{0.13\textwidth} p{0.15\textwidth} p{0.65\textwidth} }
\toprule
\textbf{LCN Category} & \textbf{Specific Aspect} & \textbf{Description} \\
\midrule

\multirow{3}{*}{\begin{tabular}[c]{l}$\mathcal{F}:$ Framing \&\\ Composition\end{tabular}}
& $\mathcal{F}_1$ Field of View & Ensuring all currently relevant active instruments and critical anatomical structures are \textit{contained within the camera frame}. \\
\cmidrule(l){2-3}
& $\mathcal{F}_2$ Centering & Keeping the primary area of surgical action and/or active instrument tips centrally located within the frame. \\
\cmidrule(l){2-3}
& $\mathcal{F}_3$ Magnification/ \newline Zoom & Using a level of zoom appropriate for the current surgical task, balancing detail with necessary contextual overview. \\
\midrule

\multirow{4}{*}{\begin{tabular}[c]{l}$\mathcal{V}:$ Visibility \&\\ Clarity\end{tabular}}
& $\mathcal{V}_1$ Instrument \newline Visibility & Ensuring instrument working ends \textit{within the frame} are clearly seen and not blocked by tissue, blood, or smoke. \\
\cmidrule(l){2-3}
& $\mathcal{V}_2$ Focus & Maintaining sharp focus, especially on the region of interest/active instruments. \\
\cmidrule(l){2-3}
& $\mathcal{V}_3$ Lighting \& \newline Exposure & Providing adequate illumination without excessive glare, deep shadows, or over/under-saturation. \\
\cmidrule(l){2-3}
& $\mathcal{V}_4$ Lens Cleanliness & Keeping the lens free from obstructions like fog, blood, smoke, condensation, or surgical debris. \\
\midrule

\multirow{2}{*}{\begin{tabular}[c]{l}$\mathcal{O}:$ Orientation \&\\ Stability\end{tabular}}
& $\mathcal{O}_1$ Horizon Alignment & Maintaining a consistent, level (horizontal) camera orientation to aid spatial understanding. \\
\cmidrule(l){2-3}
& $\mathcal{O}_2$ Image Stability & Minimizing unintentional high-frequency shaking, jitter, or vibration in the camera view. \\
\midrule

\multirow{3}{*}{\begin{tabular}[c]{l}$\mathcal{M}:$ Motion \&\\ Dynamics\end{tabular}}
& $\mathcal{M}_1$ Smoothness & Executing deliberate camera movements (panning, tilting, zooming) fluidly without abruptness or \newline jerkiness. \\
\cmidrule(l){2-3}
& $\mathcal{M}_2$ Economy of \newline Motion & Moving the camera purposefully and efficiently, minimizing unnecessary or extraneous movements. \\
\cmidrule(l){2-3}
& $\mathcal{M}_3$ Responsiveness \& \newline Anticipation & Following surgeon's actions promptly and, ideally, anticipating upcoming needs for proactive \newline positioning. \\
\midrule

\multirow{2}{*}{\begin{tabular}[c]{l}$\mathcal{S}:$ Safety \&\\ Awareness\end{tabular}}
& $\mathcal{S}_1$ Collision Avoidance & Navigating to avoid unintended contact with internal structures, other instruments, or the scope itself. \\
\cmidrule(l){2-3}
& $\mathcal{S}_2$ Contextual \newline Awareness & Maintaining a viewing distance and angle reflecting an understanding of the broader surgical \newline phase/context. \\
\bottomrule
\end{tabular}
\end{table*}

\begin{figure*}[ht!]
\centering
\includegraphics[width=0.9\textwidth]{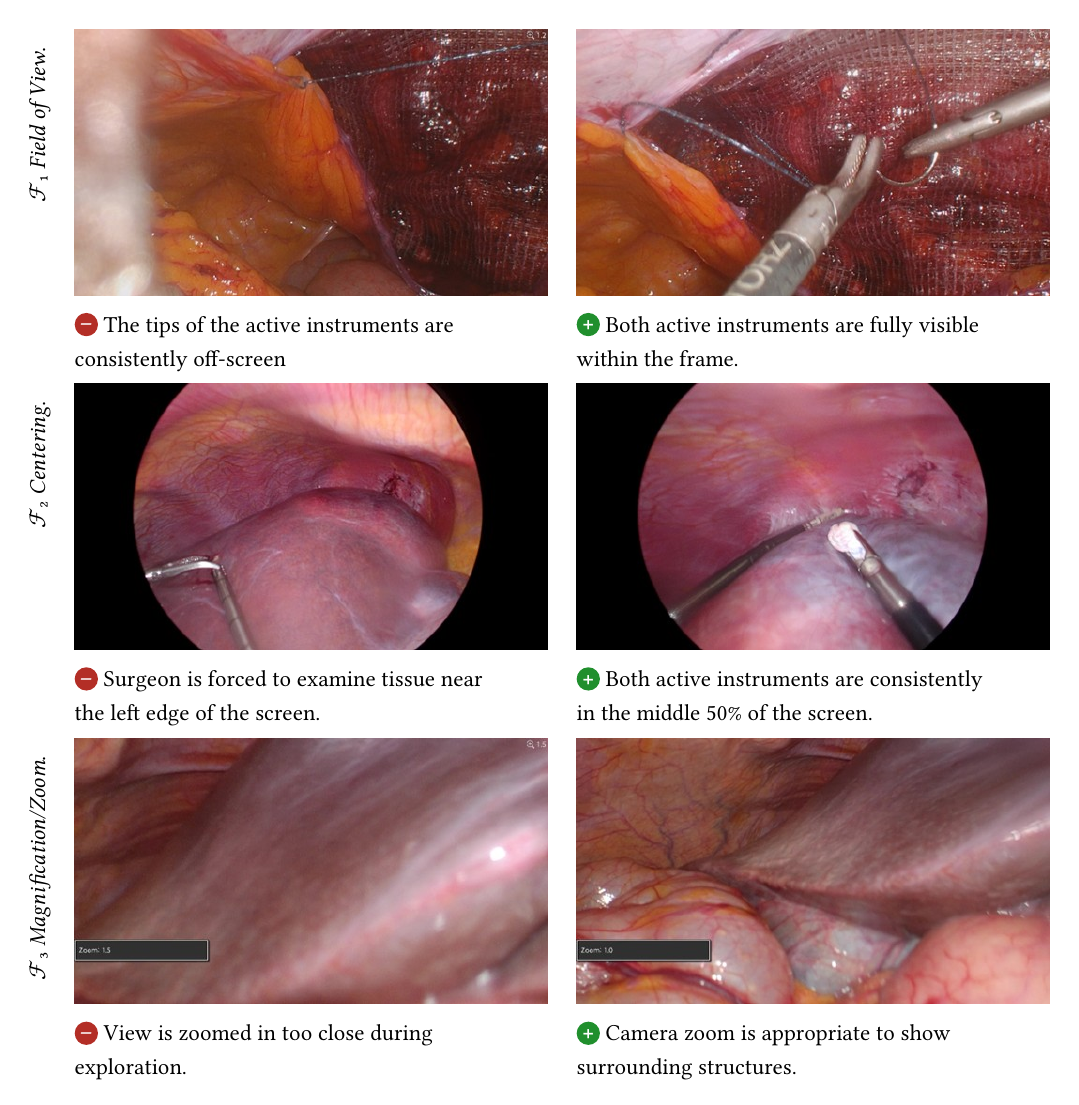}
\caption{Good and bad examples for each LCN aspect in the category \emph{Framing~\&~Composition}}\label{f.examples-framing-composition}
\end{figure*}

\subsection{Evaluation of Clinical Relevance via Surgeon Survey}

To evaluate the clinical relevance of the 14 skills, a structured online survey was conducted using LimeSurvey. The survey was distributed to practicing surgeons at the Clinic for General, Visceral, and Pediatric Surgery of the University Medical Center Göttingen (UMG). Participants were encouraged to disseminate the survey, resulting in participation from external institutions including the University Medical Center Hamburg-Eppendorf (UKE) and the University Medical Center Schleswig-Holstein (UKSH). As institutional affiliation disclosure was optional, a subset of respondents remains uncategorized (Appendix-Figure~\ref{f.demographic-distrib}).

\subsubsection{Survey Design and Content}

The questionnaire comprised three sections, designed for completion in 5--10 minutes to minimize participation obstacles. Participation was anonymous and voluntary, with all subjects providing informed consent.

\textbf{Section 1: Demographic Information.} This section collected professional background data to contextualize responses. Required information included country of practice, primary surgical specialties, post-residency experience, laparoscopic experience years, annual procedure volume, and resident training experience. Institutional affiliation was optional.

\textbf{Section 2: Importance of Navigation Aspects.} This section systematically evaluated the 14 camera navigation skills within the five subgroups. For each skill, a precise definition was provided, and participants rated its importance on a 5-point Likert scale (1 = Not at all Important to 5 = Extremely Important). Optional open-text fields allowed qualitative feedback.

\textbf{Section 3: Overall Feedback and Prioritization.} The final section employed a forced-choice question requiring participants to select the five most essential aspects for excellent LCN. Additionally, surgeons rated the overall importance of good LCN regarding four outcomes: \emph{Surgical Safety}, \emph{Surgical Efficiency}, \emph{Surgeon Comfort/Reduced Frustration}, and \emph{Effectiveness of Surgical Assistant Training}. The survey concluded with open-ended questions to identify potentially missing skills.

\subsubsection{Survey Data Analysis}

The collected data underwent quantitative assessment. While thematic analysis was planned for open-ended responses, no qualitative comments were received; consequently, analysis relied exclusively on quantitative metrics.

\textbf{Quantitative Analysis:} Descriptive statistics (median, mode, and interquartile range) were calculated for the Likert ratings. Additionally, we calculated how often experts selected each skill as a “top-five” priority to establish a ranked hierarchy of critical skills.

To prioritize skills, a dual-criteria ranking methodology classified items into hierarchical categories based on two metrics: selection count ($C$) and the Top-Two-Box (T2B) percentage. The T2B percentage represents the proportion of respondents who rated a given aspect as either ``Very Important'' (4) or ``Extremely Important'' (5) on the 5-point Likert scale.

Two independent ranked lists were generated: one (Count Rank) ordered by $C$ (top-five priority selection frequency) and the other (Likert Rank) ordered by the T2B percentage. To resolve ties in the Likert Rank, the Top-Box percentage--the proportion of respondents rating the aspect exclusively as ``Extremely Important'' (5)--was incorporated as a secondary tiebreaker. The combined rank ($R$) was then computed by summing the Likert Rank and Count Rank, followed by applying a dense ranking to these sums in ascending order. Resulting classifications were defined as follows (Appendix-Figure~\ref{f.ranking-procedure}):

\textbf{Highly Important Aspects:} Represents the intersection of both metrics. Includes aspects ranking within the top-5 of \emph{both} the selection count and average rating lists, indicating consensus on high importance regarding popularity and perceived value.

\textbf{Notably Important Aspects:} Includes aspects ranking in the top-5 of \emph{either} the priority selection list \emph{or} the average rating list, but not both. This distinction separates skills with significant but singularly defined importance from universally critical items.

\subsection{Evaluating Technological Readiness of CV Approaches}

Following taxonomy development, the technical readiness of potential quantitative approaches for the 14 skills was evaluated based on SoTA CV. This assessment focused on metrics measurable from a standard two-dimensional surgical video feed without reliance on specialized hardware. The evaluation accounted for intraoperative challenges like instrument detection, occlusion, visual artifacts (e.g., smoke, blood), and motion estimation. Technological readiness was classified into three levels:

Given the rapid evolution of CV, this section presents a narrative synthesis and expert technological readiness assessment rather than a systematic review.

\textbf{High:} Algorithms demonstrating robust clinical translation potential, strictly characterized by an accuracy or Dice similarity coefficient (DSC) of $>90\%$ on in vivo surgical datasets, coupled with the capability for real-time processing, i.e., $\ge 30$~frames per second (fps).

\textbf{Medium:} Foundational approaches exhibiting partial feasibility, defined by an accuracy or DSC of 70\%--90\%, operation below real-time temporal constraints ($<30$~fps), or empirical validation restricted solely to ex vivo or highly constrained experimental datasets.

\textbf{Low:} Methods necessitating fundamental research advances, quantitatively bound by $<70\%$ accuracy, elevated spatial error margins (e.g., root-mean-square error [RMSE] $>10$\,mm), an absence of standardized benchmarking, or an exclusive reliance on simulated environments.

Finally, survey findings \emph{(Clinical Importance)} were synthesized with the technical assessment \emph{(CV Technological Readiness)} to generate a prioritization matrix guiding future automated assessment tool development.

\section{Results}

In the following, we present the results for each of the three methodological phases, presented in Section~\ref{s2.methods}.

\subsection{Taxonomy of LCN Aspects}

Table~\ref{t.overview} outlines all identified aspects defining good LCN. For each of the 14 aspects, we created a structured definition to ensure unambiguous understanding by laparoscopic surgeons. Figure~\ref{f.examples-framing-composition} depicts good and bad examples for \emph{Framing~\&~Composition}; see Appendix-Figure~\ref{f.examples-visibility-clarity} (p.~\pageref{f.examples-visibility-clarity}) and Appendix-Table~\ref{t.examples} (p.~\pageref{t.examples}) for additional examples.

\subsection{Survey Results}

Out of 23 total survey respondents in the filtered consensus cohort, all provided complete demographic profiles. The baseline demographic data is detailed in Table~\ref{t.demographics}. Among the respondents who specified their laparoscopic surgical experience ($n=23$), 39.1\% ($n=9$) were classified as Novices ($\le 5$ years of experience), and 60.9\% ($n=14$) were classified as Experienced ($>5$ years of experience). Annual laparoscopic procedure volume was well-distributed across the cohort, with 26.1\% ($n=6$) performing 25--50 procedures per year and 13.0\% ($n=3$) performing over 200. The primary surgical specialty represented in the multi-select responses was Visceral surgery (general) (73.9\%, $n=17$), followed by Upper GI / bariatric surgery (43.5\%, $n=10$) and Hepatobiliary surgery (43.5\%, $n=10$). Training experience in LCN was predominantly reported as limited (61.1\%, $n=11$).

Given the small size of this exploratory cohort ($N=23$), statistical hypothesis testing for differences between subgroups (such as Novice vs. Experienced surgeons) was omitted to avoid overinterpreting low-powered statistical tests. Instead, differences in clinical importance ratings for the 14 LCN aspects are presented descriptively. Descriptively, both experience groups aligned closely on most aspects, reflecting a high level of consensus. For instance, Field of View was rated highly by both cohorts, with novices demonstrating a slightly higher descriptive rating (Median = 5.0, IQR = 1.0) than experienced surgeons (Median = 4.0, IQR = 0.8). Centering also showed strong endorsement in both subgroups (Novice Median = 5.0, IQR = 1.0 vs. Experienced Median = 4.0, IQR = 1.0). Ratings for motion-related aspects, such as Economy of Motion, remained stable and moderate across groups (Novice Median = 3.0, IQR = 1.0 vs. Experienced Median = 3.0, IQR = 1.0). Safety-focused metrics, such as Contextual Awareness, showed identical descriptive distributions between subgroups (Novice Median = 4.0, IQR = 0.0 vs. Experienced Median = 4.0, IQR = 0.0). These descriptive observations indicate that novices and experienced surgeons share a largely consistent perspective on LCN clinical priorities, without requiring formal statistical inference.

\begin{table}[t!]
\centering
\caption{Demographic characteristics of the survey respondents. The response frequency is denoted by $n$. Note that \emph{Primary Surgical Specialties} is a multi-select variable. Percentages are calculated based on the $N = 23$ total respondents.}
\label{t.demographics}
\small
\begin{tabular}{p{0.37\textwidth} c c}
\toprule
\textbf{Demographic Variable} & \textbf{$\boldsymbol{n}$} & \textbf{\%} \\
\midrule

\textbf{Post-Residency Surgical Experience}\\
\textit{(Valid $N = 23$, Missing $= 0$)} & & \\
\quad Pre-residency & 3 & 13.0 \\
\quad $< 5$ years & 6 & 26.1 \\
\quad 5--10 years & 8 & 34.8 \\
\quad 11--15 years & 3 & 13.0 \\
\quad 16--20 years & 1 & 4.3 \\
\quad $> 20$ years & 2 & 8.7 \\
\addlinespace

\textbf{Laparoscopic Surgical Experience}\\
\textit{(Valid $N = 23$, Missing $= 0$)} & & \\
\quad 1--5 years (Novice) & 9 & 39.1 \\
\quad 6--10 years (Experienced) & 7 & 30.4 \\
\quad 11--15 years (Experienced) & 1 & 4.3 \\
\quad $> 15$ years (Experienced) & 6 & 26.1 \\
\addlinespace

\textbf{Annual Laparoscopic Procedure Volume}\\
\textit{(Valid $N = 23$, Missing $= 0$)} & & \\
\quad $< 25$ & 5 & 21.7 \\
\quad 25--50 & 6 & 26.1 \\
\quad 51--100 & 5 & 21.7 \\
\quad 101--200 & 4 & 17.4 \\
\quad $> 200$ & 3 & 13.0 \\
\addlinespace

\textbf{Training Experience in LCN}\\
\textit{(Valid $N = 18$, Missing $= 5$)} & & \\
\quad None & 0 & 0.0 \\
\quad Limited & 11 & 61.1 \\
\quad Moderate & 4 & 22.2 \\
\quad Extensive & 3 & 16.7 \\
\addlinespace

\textbf{Primary Surgical Specialties}\\
\textit{(Total $N = 23$, Missing $= 0$)} & & \\
\quad Visceral surgery (general) & 17 & 73.9 \\
\quad Upper GI / Bariatric Surgery & 10 & 43.5 \\
\quad Hepatobiliary Surgery & 10 & 43.5 \\
\quad Colorectal Surgery & 7 & 30.4 \\
\quad Abdominal Wall / Hernia Surgery & 6 & 26.1 \\
\quad Endocrine Surgery & 3 & 13.0 \\
\quad Transplant Surgery & 3 & 13.0 \\
\quad Thoracic Surgery & 2 & 8.7 \\
\quad Pediatric Surgery & 2 & 8.7 \\
\quad Gynecology & 0 & 0.0 \\
\quad Urology & 0 & 0.0 \\
\quad Other Specialty & 0 & 0.0 \\
\bottomrule

\end{tabular}
\end{table}

Table~\ref{t.ranking} presents the relevance ranking $R$ for all LCN aspects, derived from surgeon consensus and importance ratings ($N=23$). Each aspect was rated on a five-point Likert scale ($1 = $ Not at all Important to $5 = $ Extremely Important); because Likert scale intervals cannot be assumed equal, means and standard deviations are included for reference only, rankings use ordinal statistics and top-box percentages. Two metrics inform the ranking: $C$, the frequency with which an aspect was selected in a top-5 choice question (\emph{``If you could only choose five aspects [\ldots] which five would you choose?''}), and \emph{T2B\%}, the Top-Two-Box percentage (proportion of ratings $\ge 4$). Both metrics are ranked in descending order, so that the highest values receive rank~1, yielding $r(C)$ and $r(T2B\%)$, respectively; for $r(T2B\%)$, the Top-Box percentage (proportion of ratings $= 5$) serves as a tiebreaker. The combined ranking $R$ is then defined as $r(r(C) + r(T2B\%))$. Additional descriptive statistics reported in the table include the mean~$\mu$, median~\emph{Mdn}, mode~\emph{Mo}, interquartile range~\emph{IQR}, standard deviation~$\sigma$ (lower values indicate higher agreement), and the minimum and maximum ratings \emph{min} and \emph{max}. Row coloring corresponds to the ranking procedure illustrated in Appendix-Figure~\ref{f.ranking-procedure}. Notably, the top three aspects Centering, Focus, and Field of View exhibit the lowest standard deviations.

Table~\ref{t.general-importance} details ratings of the perceived importance of good LCN for surgical outcomes, specifically \emph{safety}, \emph{efficiency}, \emph{surgeon comfort}, and \emph{assistant training}.

\newcommand*\circled[1]{\tikz[baseline=(char.base)]{\node[shape=circle,draw,inner sep=1.5pt] (char) {#1};}}

\definecolor{rowcolorRed1}{RGB}{251, 205, 205}
\definecolor{rowcolorRed2}{RGB}{253, 223, 223}
\definecolor{rowcolorOrange1}{RGB}{254, 233, 218}
\definecolor{rowcolorOrange2}{RGB}{254, 223, 200}
\definecolor{rowcolorGrey}{RGB}{230, 230, 230}

\begin{table*}[ht!]
\centering
\renewcommand{\arraystretch}{1.2} 
\setlength{\tabcolsep}{8pt}      
\caption{Ranking of the LCN aspects.}\label{t.ranking}
\begin{tabular}{l c c c c c c c c c c c c}
\toprule
\textbf{LCN Aspect} & \textbf{$\boldsymbol{R}$} & \textbf{$\boldsymbol{r(C)}$} & \textbf{$\boldsymbol{C}$} & \textbf{$\boldsymbol{r(T2B)}$} & \textbf{\textit{T2B\%}} & \textbf{$\boldsymbol{\mu}$} & \textbf{\textit{Mdn}} & \textbf{\textit{Mo}} & \textbf{\textit{IQR}} & \textbf{$\boldsymbol{\sigma}$} & \textbf{min} & \textbf{max} \\
\midrule
\rowcolor{rowcolorRed1}
$\mathcal{F}_{2}$ Centering & 1 & 1 & 19 & 2 & 100 & 4.43 & 4 & 4 & 1 & 0.5 & 4 & 5 \\
\rowcolor{rowcolorRed2}
$\mathcal{V}_{2}$ Focus & 2 & 4 & 11 & 1 & 100 & 4.65 & 5 & 5 & 1 & 0.48 & 4 & 5 \\
\rowcolor{rowcolorRed1}
$\mathcal{F}_{1}$ Field of View & 2 & 2 & 15 & 3 & 95.7 & 4.39 & 4 & 4 & 1 & 0.57 & 3 & 5 \\
\rowcolor{rowcolorOrange1}
$\mathcal{V}_{4}$ Lens Cleanliness & 3 & 6 & 9 & \circled{4} & 87 & 4.22 & 4 & 4 & 1 & 0.66 & 3 & 5 \\
\rowcolor{rowcolorOrange2}
$\mathcal{V}_{1}$ Instrument Visibility & 4 & \circled{3} & 12 & 8 & 69.6 & 3.91 & 4 & 4 & 2 & 0.93 & 2 & 5 \\
\rowcolor{rowcolorOrange1}
$\mathcal{V}_{3}$ Lighting \& Exposure & 5 & 9 & 5 & \circled{5} & 82.6 & 4.09 & 4 & 4 & 0.5 & 0.65 & 3 & 5 \\
\rowcolor{rowcolorOrange2}
$\mathcal{O}_{1}$ Horizon Alignment & 5 & \circled{5} & 10 & 9 & 69.6 & 3.91 & 4 & 4 & 1 & 0.72 & 3 & 5 \\
$\mathcal{M}_{3}$ Responsiveness \& Anticipation & 6 & 8 & 7 & 7 & 73.9 & 3.96 & 4 & 4 & 1 & 0.81 & 2 & 5 \\
\rowcolor{rowcolorGrey}
$\mathcal{O}_{2}$ Image Stability & 6 & 6 & 9 & 9 & 69.6 & 3.91 & 4 & 4 & 1 & 0.72 & 3 & 5 \\
$\mathcal{S}_{2}$ Contextual Awareness & 7 & 10 & 3 & 6 & 78.3 & 4 & 4 & 4 & 0 & 0.66 & 3 & 5 \\
\rowcolor{rowcolorGrey}
$\mathcal{S}_{1}$ Collision Avoidance & 8 & 12 & 2 & 11 & 65.2 & 3.7 & 4 & 4 & 1 & 0.69 & 2 & 5 \\
$\mathcal{M}_{2}$ Economy of Motion & 8 & 10 & 3 & 13 & 43.5 & 3.43 & 3 & 3 & 1 & 0.77 & 2 & 5 \\
\rowcolor{rowcolorGrey}
$\mathcal{F}_{3}$ Magnification/Zoom & 9 & 12 & 2 & 12 & 52.2 & 3.48 & 4 & 4 & 1 & 0.88 & 1 & 5 \\
$\mathcal{M}_{1}$ Smoothness & 1 & 14 & 1 & 14 & 34.8 & 3.35 & 3 & 3 & 1 & 0.63 & 2 & 5 \\
\bottomrule
\end{tabular}
\end{table*}

\begin{table*}[ht!]
\centering
\renewcommand{\arraystretch}{1.25} 
\setlength{\tabcolsep}{8pt}      
\caption{Importance of good LCN to factors influencing surgical effectiveness, as rated by respondents.}
\label{t.general-importance}
\begin{tabular}{l c c c c c c c c c}
\toprule
\textbf{Factors Influencing Surgical Effectiveness} & \textbf{\textit{T2B\%}} & \textbf{\textit{n}} & \textbf{$\boldsymbol{\mu}$} & \textbf{\textit{Mdn}} & \textbf{\textit{Mo}} & \textbf{\textit{IQR}} & \textbf{$\boldsymbol{\sigma}$} & \textbf{\textit{min}} & \textbf{\textit{max}} \\
\midrule
Surgical Safety & 91.3 & 23 & 4.13 & 4 & 4 & 0.5 & 0.68 & 2 & 5 \\
Surgical Efficiency(e.g., reducing operative time) & 78.3 & 23 & 4.26 & 4 & 5 & 1 & 0.79 & 3 & 5 \\
Surgical Comfort / Reduced Frustration & 82.6 & 23 & 4.26 & 4 & 5 & 1 & 0.74 & 3 & 5 \\
Effectiveness of Surgical Assistant Training & 65.2 & 23 & 3.83 & 4 & 4 & 1 & 0.82 & 2 & 5 \\
\bottomrule
\end{tabular}
\end{table*}

\subsection{CV Approaches}

We evaluate the technological readiness of CV pipelines for quantifying each LCN aspect based on current SoTA video analysis capabilities. Applying the strict, rule-based categorization thresholds detailed in Section~\ref{s2.methods}, we classify the clinical translation potential of each method as High, Medium, or Low based on their accuracy thresholds and real-time processing constraints. Table~\ref{t.readiness_assessment} synthesizes the core CV methods, specific performance metrics, key citations, and final readiness designations for all 14 LCN aspects.

\begin{table*}[htbp]
\centering
\caption{Technical Readiness Assessment by LCN Aspect.}
\label{t.readiness_assessment}
\small
\renewcommand{\arraystretch}{2}
\begin{tabular}{@{} p{0.15\textwidth} p{0.18\textwidth} p{0.40\textwidth} c p{0.10\textwidth} @{}}
\toprule
\textbf{LCN Aspect} & \textbf{Core CV Method} & \textbf{Specific Performance Metrics} & \textbf{Readiness} & \textbf{Citations} \\
\midrule
$\mathcal{F}_1$ Field of View & Multi-object detection \newline \& spatio-temporal tracking & \textbullet~93\% tool localization accuracy \newline \textbullet~33.19 fps processing (\textit{in vivo}) & High & \cite{Zhao.2019,Qiu.2019,Wang.2022,Nwoye.2023} \\

$\mathcal{F}_2$ Centering & Instance segmentation \newline \& keypoint detection & \textbullet~DRR-Net: Dice 96.27\%, mIoU 92.82\% \newline \textbullet~YOLOv8+ByteTrack: F1 92.0\% @ 45 fps \newline \textbullet~ART-Net: 100.0\% acc, 9.3px error & High & \cite{Lavanchy.2021,Yang.2022,Myo.2024,Hasan.2021} \\

$\mathcal{F}_3$ Magnification/ \newline Zoom & Segmentation pixel area \newline \& surgical phase models & \textbullet~EndoNet: 81.0\% AP @ 5 fps (0.2s) \newline \textbullet~LoViT: 92.4\% (Cholec80), 81.4\% (\textit{in vivo}) @ 62.0 ms (\~16 fps) \newline \textbullet~Trans-SVNet: 78.3\% @ 100 fps & Medium & \cite{Liao.2025,Twinanda.2017,Liu.2025} \\

$\mathcal{V}_1$ Instrument \newline Visibility & Semantic segmentation \newline (instruments vs. occlusions) & \textbullet~CNNs: Dice 81.89\% @ 15-21 fps \newline \textbullet~Spatio-temporal: mIoU 54\%--69.38\% \newline \textbullet~Dynamic occlusion SSIM: 0.80--0.90 @ 7-27 fps & Medium & \cite{Marullo.2023,Grammatikopoulou.2024,Xia.2025} \\

$\mathcal{V}_2$ Focus & Image quality assessment / \newline Cascade classifiers & \textbullet~Generic models: PLCC 0.94-0.97 \newline @ 1.55s (BIBLE) - 12s (MVV) \newline \textbullet~Cascade (LVQ): 96.55\% distortion acc, \newline 99.67\% defocus acc @ 37 fps & High & \cite{Li.2016,Zhang.2018,Pang.2016,Khan.2022,Belmokeddem.2025} \\

$\mathcal{V}_3$ Lighting \& \newline Exposure & Pixel intensity histograms \& reflection detection & \textbullet~Baseline \textit{ex vivo}: 97.75\% under-exposed, 89.11\% over-exposed \newline \textbullet~Advanced \textit{in vivo}: 99.68\% acc, >90\% precision,\newline >95\% sensitivity @ 30-117 fps (<33 ms) & High & \cite{Ali.2021,Akbari.2018,Oh.2007,Tchoulack.2008,Saad.2020,Chikkerur.2011} \\

$\mathcal{V}_4$ Lens Cleanliness & Full-frame video distortion classification & \textbullet~SVMs: 96.55\%--100\% smoke/blur acc @ 37 fps \newline \textbullet~ResNet50: 100\% smoke/noise acc, \newline >99.8\% illumination/blur acc & High & \cite{Sharib.2020,Sharib.2021,Belmokeddem.2024,Belmokeddem.2025,Belmokeddem.2025.2} \\

$\mathcal{O}_1$ Horizon Alignment & Instrument axis detection via segmentation masks & \textbullet~Direct camera ATE > 10 mm \newline \textbullet~Proxy tool segmentation: Dice > 90\% @ $\ge 30$ fps & High & \cite{Hayoz.2023,Yang.2022,Zhao.2019} \\

$\mathcal{O}_2$ Image Stability & 6-DoF trajectory via visual SLAM + high-pass filter & \textbullet~Traditional models: 4 fps \newline \textbullet~Modern SLAM: up to 100 fps, 91.54\% tracking success, \newline sub-mm ATE $0.32 \pm 0.09$ mm & High & \cite{Ling.2019,Campos.2021,Wang.2024,Elvira.2024} \\

$\mathcal{M}_1$ Smoothness & Camera trajectory tracking \& motion derivatives & \textbullet~ORB-SLAM3: 91.54\% success, \newline ATE $0.32 \pm 0.09$ mm @ 30-60 fps \newline \textbullet~BDIS-SLAM: >30 Hz \newline \textbullet~EndoGSLAM: ATE $0.34 \pm 0.21$ mm @ >100 fps & High & \cite{Campos.2021,Elvira.2024,Wang.2024,Song.2024} \\

$\mathcal{M}_2$ Economy of \newline Motion & Trajectory integration \& \newline context movement detection & \textbullet~Skill prediction accuracy 83\% \newline \textbullet~inherent dependency on surgical phase recognition models (70\%--90\% accuracy tier) & Medium & \cite{Myo.2024,Campos.2021,Lee.2020} \\

$\mathcal{M}_3$ Responsiveness \& \newline Anticipation & Predictive visual control \& trajectory forecasting & \textbullet~Control MAE: 91.0px (14.2\%) \newline \textbullet~Landmark F1: 53.0\% \newline \textbullet~Forecast errors: $47^\circ$, 0.2 length \newline \textbullet~Coverage: 31.0\%--45.8\% & Low & \cite{Deng.2025,Bieck.2020,Sangalli.2026} \\

$\mathcal{S}_1$ Collision Avoidance & Monocular depth estimation \& safety area tracking & \textbullet~Monodepth2 RMSE: 16.406 mm (>10 mm threshold) \newline \textbullet~LT-SAT: $\simeq 1.60$ fps, F-measure drops to 0.52 (52\%), \newline 8 frame recovery & Low & \cite{Li.2026,Penza.2018} \\

$\mathcal{S}_2$ Contextual \newline Awareness & Surgical phase \& action triplet recognition & \textbullet~Phase Jaccard: 77.8\% @ 23.8 fps \newline \textbullet~Triplet mAP$_{\text{IVT}}$: 29.9\% (inst-verb 39.4\%, inst-target 36.9\%) & Low & \cite{Tao.2023,Nwoye.2022} \\
\bottomrule
\end{tabular}
\end{table*}

Within \emph{Framing \& Composition}, foundational geometric tasks are highly mature. Real-time multi-object detection and dynamic tracking readily clear clinical thresholds, rendering the continuous measurement of Field of View and Centering highly actionable \cite{Zhao.2019,Qiu.2019,Wang.2022,Nwoye.2023,Yang.2022,Myo.2024,Hasan.2021}. By utilizing instrument tips as reliable proxies for the surgical action centroid, current frameworks achieve near-perfect tool localization and semantic tracking at speeds comfortably exceeding 30~fps \cite{Yang.2022,Myo.2024,Hasan.2021}. Conversely, measuring the appropriateness of Magnification/Zoom remains a prominent bottleneck. While tracking pixel area changes is computationally trivial, translating these shifts into contextual appropriateness inherently depends on surgical phase recognition models \cite{Liao.2025}. Because these necessary contextual models largely fall into the 70\%--90\% accuracy tier and struggle to process full-length procedures in real-time, context-aware zoom assessment is restricted to medium readiness \cite{Twinanda.2017,Liu.2025}.

Assessing \emph{Visibility \& Clarity} presents a dichotomy between optical quality and contextual occlusion. Classical image quality assessment—when adapted for \textit{in vivo} surgical datasets via cascade classifiers and support vector machines—quantifies Focus, Lighting \& Exposure, and Lens Cleanliness with extraordinary precision \cite{Belmokeddem.2025,Akbari.2018,Oh.2007,Tchoulack.2008,Belmokeddem.2024,Belmokeddem.2025.2}. These domain-specific models easily classify blur, illumination extremes, and artifacts like smoke at high real-time speeds \cite{Belmokeddem.2025,Akbari.2018,Oh.2007,Tchoulack.2008,Belmokeddem.2024,Belmokeddem.2025.2}. However, Instrument Visibility presents a unique challenge. While semantic segmentation is well-researched, models fail to consistently distinguish between intentional clinical occlusion (e.g., actively grasping tissue) and unintentional view-obstructing artifacts without falling below real-time operational constraints \cite{Marullo.2023,Xia.2025}. As spatio-temporal segmentation models max out at medium-tier overlap metrics, robustly inferring clinical intent from dynamic occlusions limits this aspect to medium readiness \cite{Grammatikopoulou.2024,Xia.2025}.

For \emph{Orientation \& Stability}, modern advancements in spatial tracking have circumvented historical bottlenecks. Traditional direct camera pose estimation fails to establish absolute anatomical horizons due to spatial constraints, violating our high-readiness error thresholds \cite{Hayoz.2023}. However, by substituting true gravitational horizons with the dominant angle of segmented instrument shafts, Horizon Alignment leverages the robust accuracy of tool tracking to achieve high readiness \cite{Yang.2022,Zhao.2019}. Similarly, assessing Image Stability is now highly tractable thanks to modern visual simultaneous localization and mapping (SLAM) systems \cite{Campos.2021,Wang.2024,Elvira.2024}. By maintaining sub-millimeter trajectory precision at up to 100~fps, these systems provide the exact spatial data required to mathematically isolate and quantify high-frequency camera jitter without relying on computationally expensive dense optical flow fields \cite{Wang.2024,Elvira.2024}.

The underlying SLAM trajectories that solve stability also unlock \emph{Motion \& Dynamics}. Smoothness is computationally straightforward to evaluate by calculating velocity, acceleration, and jerk magnitude derivatives directly from the estimated 6-DoF position-over-time series \cite{Campos.2021,Elvira.2024,Wang.2024,Song.2024}. Yet, as assessment shifts from tracking simple kinematics to evaluating surgical workflow, technical readiness drops. Quantifying Economy of Motion demands interpreting efficiency—such as distinguishing an unintentional drift from a proactive repositioning—which relies on medium-readiness phase models \cite{Myo.2024,Campos.2021,Lee.2020}. Responsiveness \& Anticipation is the most severely limited, as trajectory forecasting and predictive visual control architectures struggle with large spatial-temporal error rates (e.g., $47^\circ$ angle prediction errors) and low coverage rates, failing to reliably predict where the surgeon will look next \cite{Deng.2025,Bieck.2020,Sangalli.2026}.

Finally, \emph{Safety \& Awareness} encapsulates the highest semantic levels of camera guidance and remains strictly at low technical readiness. Proactive Collision Avoidance is impaired by current monocular depth estimation models, which generate high spatial errors (RMSE $> 10$~mm) in deformable, low-texture abdominal environments, alongside safety-area tracking frameworks that are too slow ($\simeq 1.60$~fps) to intervene intraoperatively \cite{Li.2026,Penza.2018}. Contextual Awareness faces even steeper hurdles; while baseline surgical phase recognition approaches medium readiness, true contextual navigation requires mapping complex instrument-verb-target interactions \cite{Tao.2023}. Current action triplet recognition architectures yield average precision scores well beneath clinical utility, confirming that comprehensive semantic understanding of the surgical field is an unsolved research frontier rather than an implementable CV tool \cite{Nwoye.2022}.

\begin{figure}[t]
\centering
\includegraphics[width=0.5\textwidth]{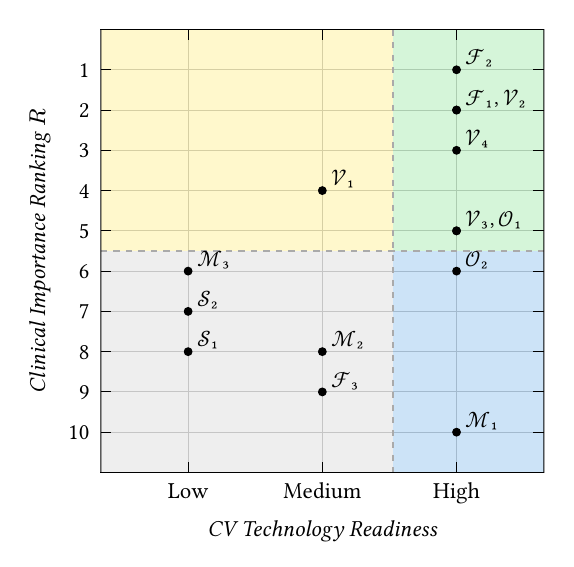}
\caption{The four-quadrant matrix of CV Technological Readiness versus Clinical Importance Ranking.}\label{f.feasibility-matrix}
\end{figure}

\subsubsection{The Clinical Importance Ranking vs.\ CV Technological Readiness Matrix}

Figure~\ref{f.feasibility-matrix} shows the correlation between the clinical importance and the technological readiness for each of the 14 aspects on a four-quadrant matrix, whose X-axis represents CV Technological Readiness (High, Medium, Low) and whose Y-axis represents Clinical Importance (defined by the ranking $R$ of all LCN aspects). The matrix is divided into four quadrants. The green quadrant (High Importance, High Readiness) contains \emph{priority targets}, aspects that are both clinically critical and technically within reach, such as \emph{Field of View~($\mathcal{F}_1$)}, \emph{Focus~($\mathcal{V}_2$)}, and \emph{Centering~($\mathcal{F}_2$)}. The yellow quadrant (High Importance, Low/Medium Readiness) highlights \emph{key research challenges}: aspects that are clinically critical but require more advanced CV approaches, including \emph{Lens Cleanliness~($\mathcal{V}_4$)}. The blue quadrant (Lower Importance, High Readiness) covers \emph{opportunistic approaches}, aspects that are easy to measure but less critical to the surgeon, and can therefore be included without being the primary focus. Finally, the gray quadrant (Lower Importance, Low Readiness) identifies \emph{low priority} aspects that are both less critical and technically hard to measure, and should therefore be deprioritized.

\section{Discussion}

\subsection{Principal Findings}

This study establishes a consensus-driven framework for automated LCN assessment. Through a systematic taxonomy of 14 distinct skills and a structured survey of 23 practicing surgeons, we identified aspects critical to surgical outcomes and evaluated their technical readiness using current CV capabilities. A primary finding is that skills surgeons deem most essential--Field of View, Centering, and Focus--are also among the most technically feasible to measure automatically. This convergence appears in the high-readiness quadrant of our clinical importance versus CV technological readiness matrix (Figure~\ref{f.feasibility-matrix}), identifying these aspects as immediate targets for automated assessment tool development.

The framework's value lies in data-driven prioritization. Rather than attempting to automate all LCN aspects simultaneously, our matrix provides a clear roadmap directing development resources toward skills that are both clinically impactful and computationally tractable. Specifically, foundational aspects such as maintaining instruments within the frame, keeping surgical action centered, and ensuring sharp focus on the region of interest emerged with highest clinical importance scores and lowest variance among surgeon ratings. These aspects form the non-negotiable baseline of competent camera work. Critically, underlying CV tasks--instrument detection and tracking, tip localization, and sharpness quantification--are well-established, robust, and readily implementable using existing open-source models and datasets.

Importance ratings support our taxonomy's relevance. Surgeons consistently rated good LCN extremely important for surgical safety, efficiency, and comfort (Table~\ref{t.general-importance}). This underscores that poor camera navigation is not merely an inconvenience but a significant contributor to operative risk and inefficiency, as documented in studies linking suboptimal visualization to prolonged procedure times and increased complication rates \cite{Nilsson.2017,Huettl.2020,Ameerah.2025,Dhingra.2025,Lacki.2025}. Our work moves beyond this general observation by quantifying which specific navigation skills matter most, enabling targeted interventions.

\subsection{Comparison to Literature}

Our approach addresses a fundamental gap in surgical training and assessment literature. Existing manual rating instruments, such as SALAS and OSA-CNS, rely on expert observers to score LCN quality \cite{Huber.2018,Nilsson.2017}. While providing valuable insights into skills such as centering and horizon alignment, these tools are inherently limited by their dependence on human raters. Although studies demonstrate high reliability under controlled validation settings--such as an Intraclass Correlation Coefficient of 0.866 and a Cronbach's alpha $> 0.7$ for the SALAS total score--manual scoring remains time-intensive, difficult to scale, and requires significant prior rater training. More critically, these frameworks were designed for human interpretation and do not decompose camera navigation into discrete, measurable features lending themselves to computational analysis.

In contrast, our work provides a bottom-up, computational blueprint. We explicitly define good LCN at a granular level, validate these definitions with practicing surgeons, and map each skill to specific CV techniques. This approach differs from prior work in two ways. First, our taxonomy is designed with automation in mind from the outset. Each of the 14 aspects is defined not as a holistic impression but as a measurable quantity--whether normalized distance of instrument tips from frame center, sharpness metric variance, or camera trajectory jerk magnitude. Second, by conducting a formal technological readiness assessment, we acknowledge that not all clinically important skills are equally amenable to current CV capabilities, preventing the pitfall of proposing theoretically desirable but computationally intractable metrics.

Our readiness analysis also reveals where the field must advance. Skills such as Responsiveness and Anticipation and Contextual Awareness were rated as moderately important but classified as low-readiness due to reliance on predictive modeling and high-level semantic understanding of surgical workflow. These aspects represent the frontier of intelligent surgical assistance, where systems perceive current states, infer intent, and anticipate actions. While recent work on surgical phase recognition and action anticipation progressed \cite{Twinanda.2017,Liu.2025,Sangalli.2026,Deng.2025}, these capabilities remain research challenges rather than deployable solutions. Our matrix makes this distinction explicit, separating what can be built today from what requires further methodological development.

The matrix also highlights an opportunistic middle ground: aspects with high readiness but moderate clinical importance, such as Lighting and Exposure and Image Stability. While not top priorities, these skills can be measured reliably using classical image processing techniques (histogram analysis, motion residual quantification). Including these metrics in an automated assessment tool incurs minimal additional computational cost and provides a more comprehensive profile of camera navigation quality.

Furthermore, it is critical to recognize the interactions within this roadmap. Computer vision development for these assessment metrics is not isolated. Developing robust instrument segmentation, for example--which is a prerequisite for measuring Centering--inherently unlocks the foundational spatial data needed for quantifying the Field of View and Economy of Motion. Understanding these interdependencies allows for more strategic prioritization of development resources, as solving fundamental visual tasks provides the scaffolding for multiple high-level LCN metrics.

\subsection{Limitations and Future Work}

This study has limitations warranting consideration. First, clinical evaluation involved 23 surgeons, mostly from the University Medical Center in Göttingen. While this cohort provided consistent ratings, particularly for highest-priority skills (evidenced by low standard deviations), the generalizability of our importance rankings to other surgical specialties, geographic regions, or experience levels remains to be established. Future work should further validate this taxonomy with a larger, more diverse international cohort to ensure prioritization robustness.

Second, our technological readiness assessment relies on literature review and expert judgment of current CV capabilities. While we identified robust methods for high-readiness aspects and acknowledged challenges for low-readiness ones, true pipeline performance in real surgical video data--with inherent variability in lighting, smoke, blood, and instrument types--requires validation through empirical implementation. Some high-readiness aspects may encounter unforeseen difficulties in deployment, while others rated as medium-readiness may prove more tractable than anticipated with recent algorithmic advances.

Third, our taxonomy, while detailed, may not be exhaustive. The survey did not elicit qualitative comments suggesting missing aspects, indicating the 14 skills captured participating surgeons' core concerns. However, as automated systems are deployed and surgeons interact with real-time feedback, additional nuances or higher-order skills may emerge warranting inclusion in future framework iterations.

The immediate next step is developing prototypes for individual, high-relevance aspects before fusing them into a comprehensive system. We propose creating targeted assessment modules for the top-right quadrant of our matrix: Field of View, Centering, and Focus. These three aspects are clinically critical and technically achievable using existing instrument detection models, tip localization algorithms, and sharpness metrics. A minimum viable solution implementing these metrics would provide quantitative feedback and serve as a foundation for automation. Validation in simulated and live surgical environments is essential to assess accuracy, real-time performance, and educational impact.

Beyond immediate implementation, this framework opens pathways for intelligent surgical assistance systems. Once reliable metrics for high-priority aspects are established, they can be integrated into robotic camera holders or augmented reality displays to provide real-time guidance or semi-autonomous camera control. For example, systems continuously monitoring centering and field of view could issue alerts when instruments drift toward frame edges or autonomously adjust positions. Such systems would augment, not replace, human assistants, reducing cognitive load and improving consistency.

Finally, this work highlights the broader need for interdisciplinary collaboration in surgical technology development. Automated assessment tool success depends on aligning clinical priorities with computational readiness--a balance achievable only through sustained dialogue between surgeons and CV researchers. Our matrix serves as a model for this collaboration, providing a shared language and visual representation enabling both communities to identify high-impact opportunities and allocate resources effectively.

\section{Conclusion}

In conclusion, this study provides the foundational blueprint for automated LCN assessment that has been lacking in the field. By defining a detailed taxonomy, evaluating it with practicing surgeons, and mapping it to CV capabilities, we have created a clear, actionable roadmap for development. The convergence of clinical importance and technical readiness for foundational skills such as Field of View, Centering, and Focus demonstrates that impactful automation is not only desirable but achievable with current technology. This framework will enable the creation of scalable, automated feedback systems that can accelerate surgical training, improve operative safety, and pave the way for intelligent assistance tools that enhance the capabilities of surgical teams.

\bmsection*{Author contributions}

Amir Ebrahimzadeh was responsible for the majority of the research activities, including conceptualization, data curation, methodology, investigation, software development, visualization, formal analysis and writing (original draft). Nazila Esmaeili contributed to writing (review and editing) through feedback and revision of the manuscript. Michael Ghadimi contributed resources by providing the necessary research infrastructure and assisting with participant recruitment for the survey. Jannis Hagenah was responsible for supervision of the project and also contributed to writing (review and editing).

\bmsection*{Acknowledgments}

This study was supported by the Ministry of Science and Culture (Ministerium für Wissenschaft und Kultur - MWK) of Lower Saxony (grant number ZN4094). Furthermore, we acknowledge support by the Open Access Publication Funds of the Göttingen University.
\bmsection*{Conflict of interest}

The authors declare no potential conflict of interests.

\pagebreak

\bibliography{ref}

\appendix

Appendix starts on the next page.

\pagebreak

\begin{figure*}[t]
\centering
\includegraphics[width=0.9\textwidth]{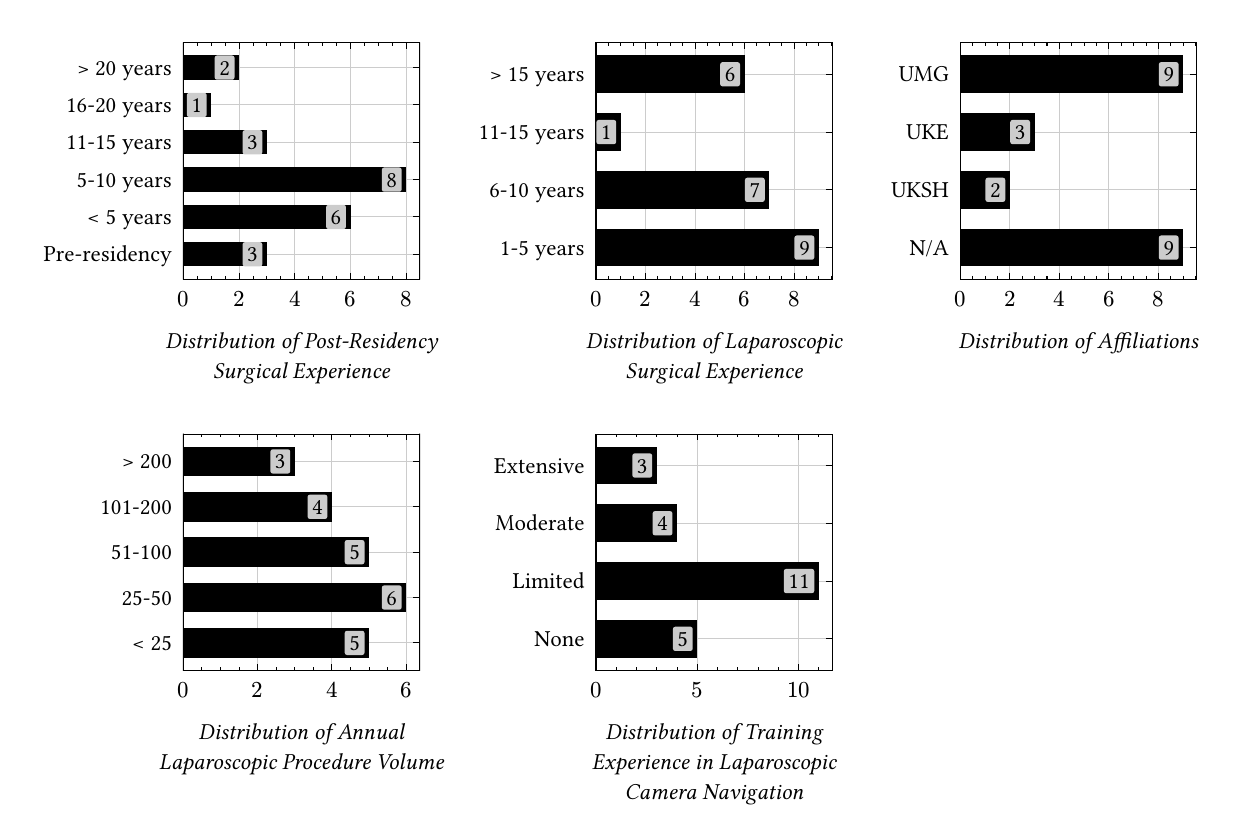}
\caption{Distribution of survey responses regarding experience levels and affiliations. The majority of participants were affiliated with the University Medical Center Göttingen (UMG). Some participants were affiliated with the University Medical Center Hamburg-Eppendorf (UKE) and the University Medical Center Schleswig-Holstein (UKSH). N/A represents all survey participants who did not provide information about their affiliation.}\label{f.demographic-distrib}
\end{figure*}

\begin{figure*}[t]
\centering
\includegraphics[width=0.8\textwidth]{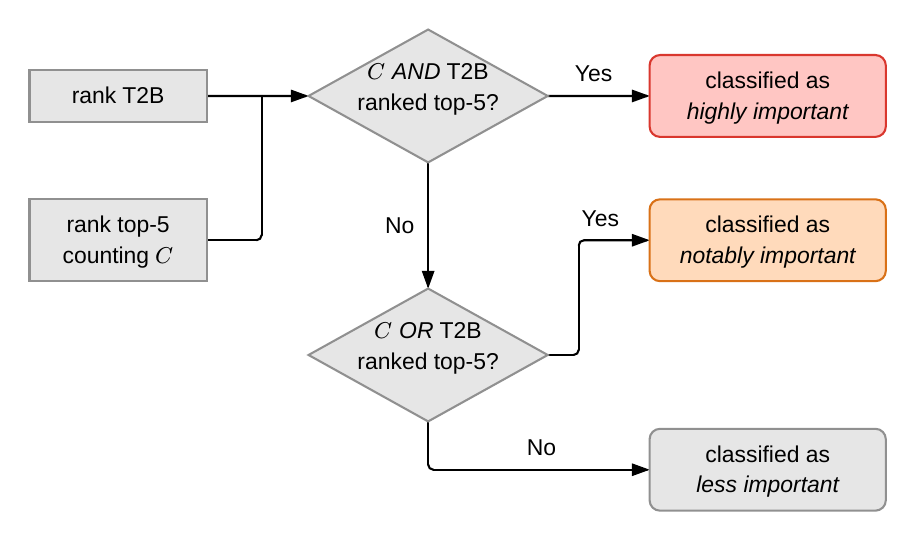}
\caption{Ranking procedure of the survey importance ratings.}
\label{f.ranking-procedure}
\end{figure*}

\begin{figure*}[t]
\centering
\includegraphics[width=0.8\textwidth]{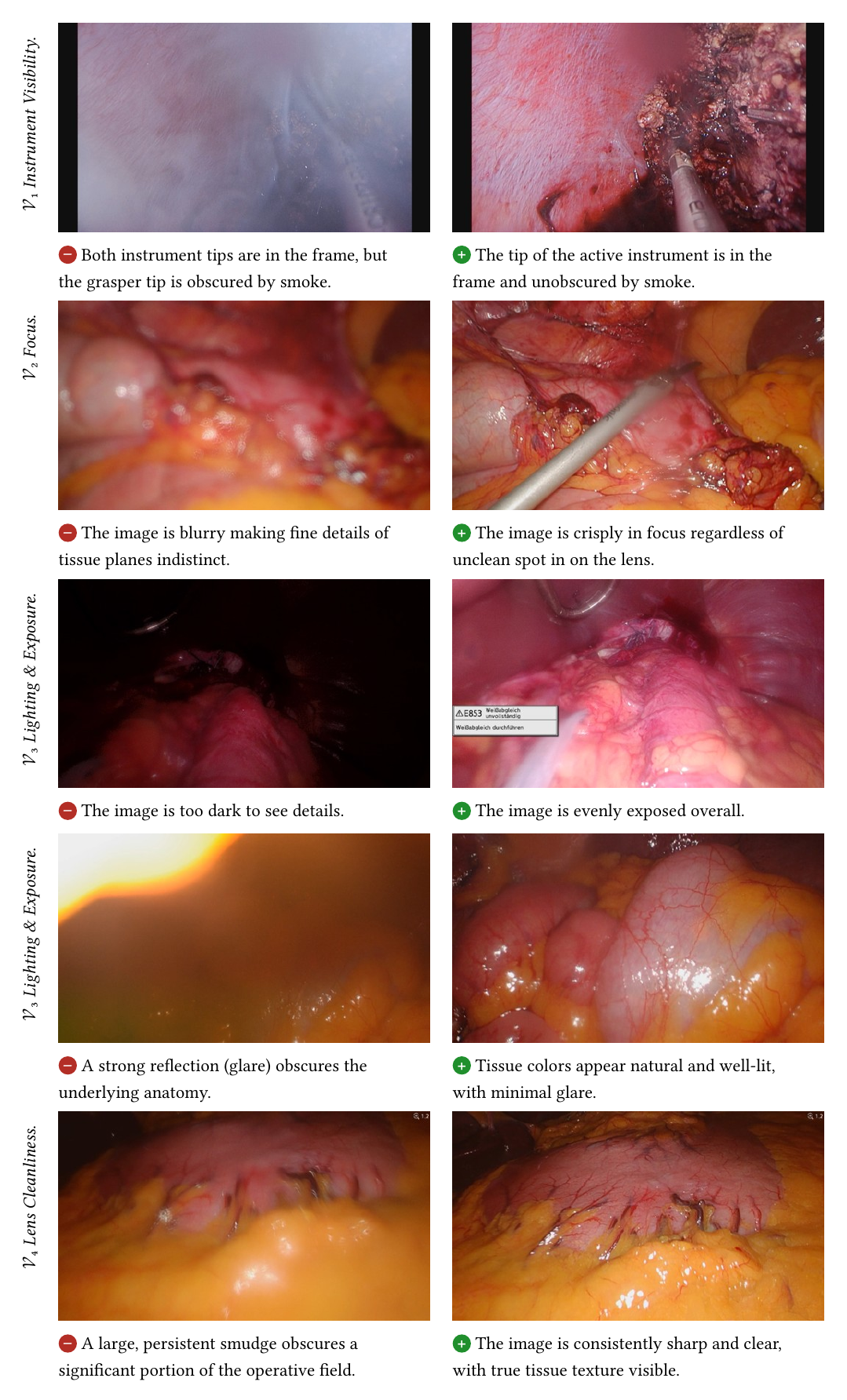}
\caption{Good and bad examples for each LCN aspect in the category \emph{Visibility~\&~Clarity}}\label{f.examples-visibility-clarity}
\end{figure*}

\begin{table*}[t]
\renewcommand{\arraystretch}{1.85} 
\caption{Good and bad examples LCN aspects from the categories \emph{Orientation~\&~Stability}, \emph{Motion~\&~Dynamics}, and \emph{Safety~\&~Awareness}.}
\label{t.examples}
\begin{tabular}{p{0.17\textwidth} p{0.38\textwidth} p{0.38\textwidth}}
\toprule
\textbf{LCN Aspect} & \textbf{Bad Example} & \textbf{Good Example} \\
\midrule
$\mathcal{O}_{1}$ ~Horizon Alignment & The camera view is consistently tilted 45 degrees, making anatomical planes and instrument angles confusing to judge. & The view remains level, mimicking a natural perspective, even as the camera pans and tilts to follow the action. \\
$\mathcal{O}_{2}$ ~Image Stability & The image is constantly jittery or shaky, making it difficult for the surgeon to perform precise tasks and causing eye strain. & The image is rock-steady during a precise dissection, \newline allowing the surgeon to make fine movements confidently. \\
$\mathcal{M}_{1}$ ~Smoothness & Camera movements are sudden and jerky, often overshooting the target or causing a disorienting "whip-pan" effect. & When the surgeon moves to a new area, the camera pans gracefully and stops precisely on target without overshooting. \\
$\mathcal{M}_{2}$ ~Economy of Motion & The camera operator is constantly making small, fidgety \newline adjustments or drifts the camera aimlessly when no surgical action requires it. & The camera operator makes only essential movements, \newline keeping the view stable on the action unless a clear need to reposition arises. \\
$\mathcal{M}_{3}$ ~Responsiveness\newline \& Anticipation & The camera consistently lags behind the surgeon's \newline movements, forcing the surgeon to wait or work off-center temporarily. & As the surgeon's instrument moves towards a new structure, the camera is already adjusting to frame that structure. \\
$\mathcal{S}_{1}$ ~Collision Avoidance & The camera lens bumps into an organ, causing a sudden jarring of the image and potential minor trauma or bleeding. & The camera moves smoothly within the abdominal cavity, maintaining a safe distance from organs and instruments. \\
$\mathcal{S}_{2}$ ~Contextual Awareness & The camera remains extremely zoomed in on a small area even when the surgeon is trying to identify larger anatomical landmarks for orientation. & During a wider exploration phase, the camera provides a broader view; during dissection of a small vessel, it is \newline appropriately zoomed in. \\
\bottomrule
\end{tabular}
\end{table*}

\begin{figure*}[hb!]
\centering
\includegraphics[width=0.5\textwidth]{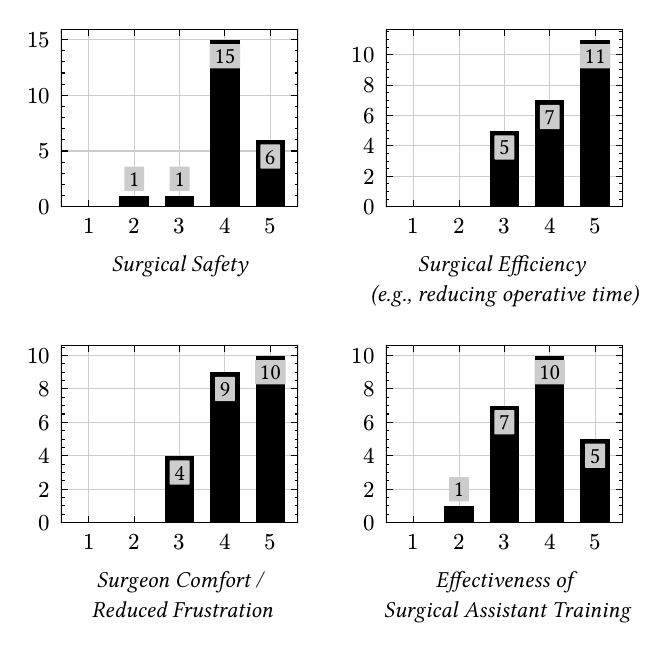}
\caption{Distribution of survey responses regarding how LCN influences general factors of surgical effectiveness.}\label{f.general-importance-rating-distrib}
\end{figure*}

\begin{figure*}[t]
\centering
\includegraphics[width=0.8\textwidth]{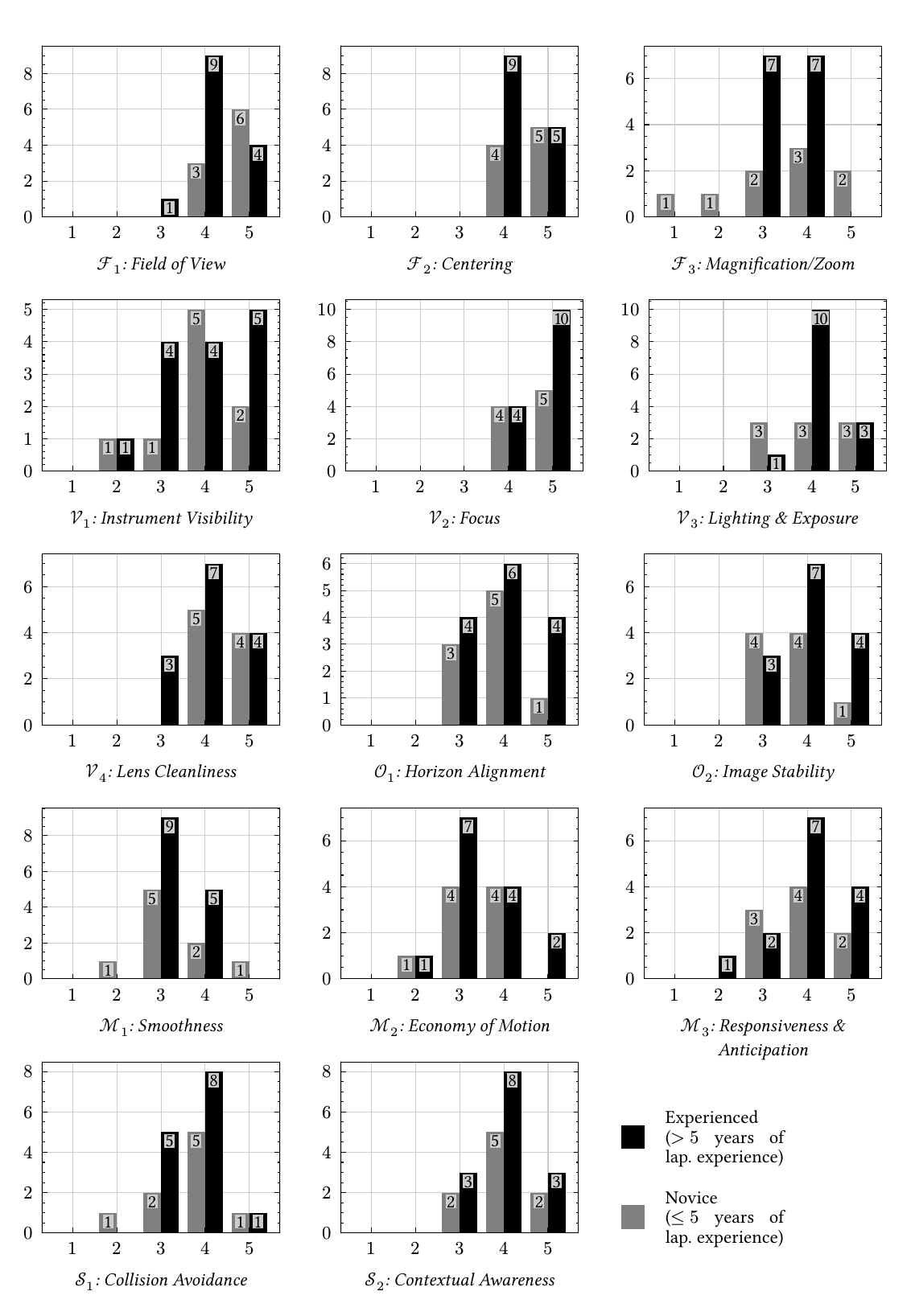}
\caption{Distribution of survey responses regarding importance rating for each LCN aspect.}\label{f.importance-rating-distrib}
\end{figure*}

\begin{figure*}[htp!]
\centering
\includegraphics[width=0.95\textwidth, page=1]{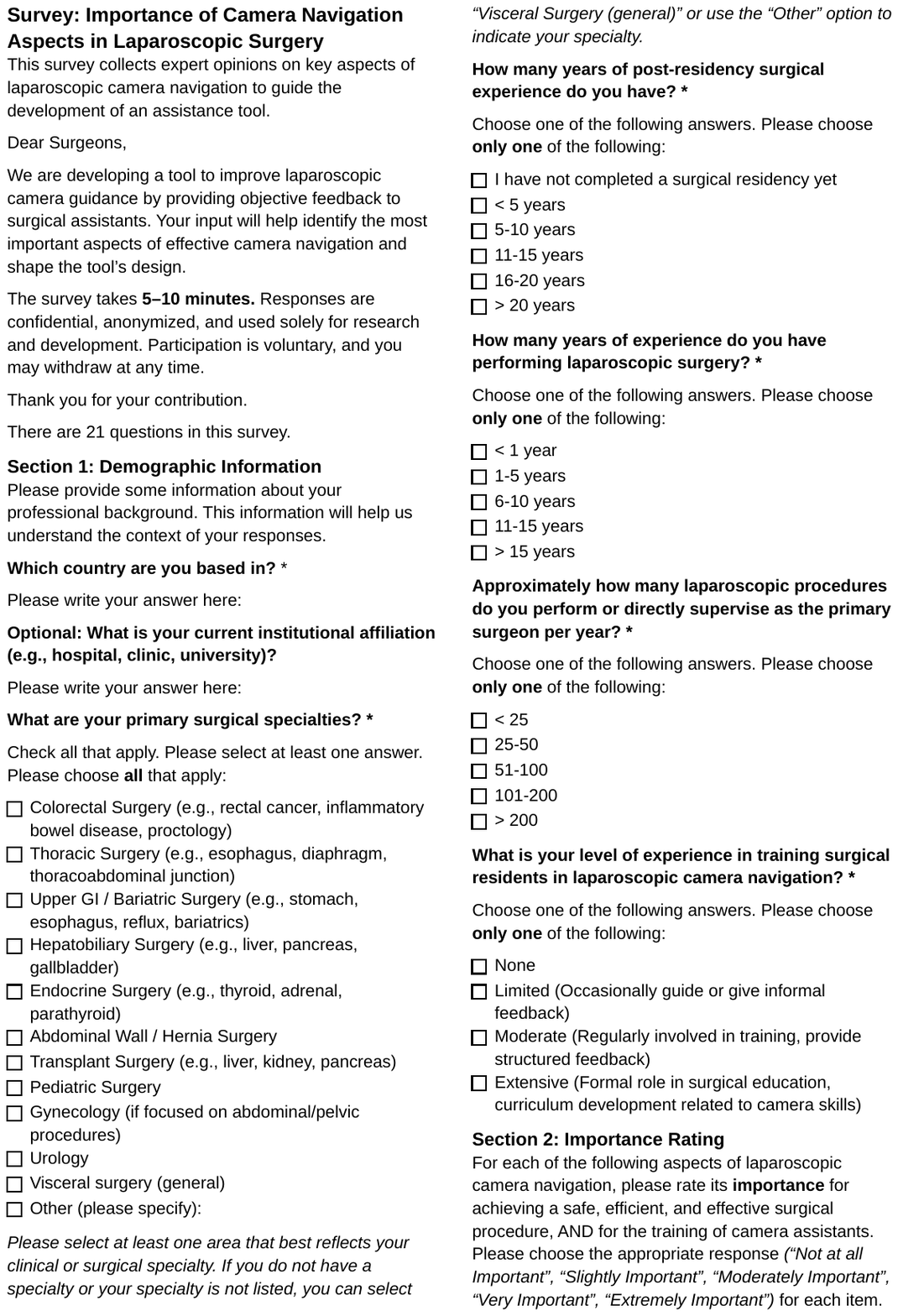}
\caption{Written form of the LimeSurvey online questionnaire, including all original sections and questions (page 1 of 2).}\label{survey.p1}
\end{figure*}

\begin{figure*}[htp!]
\centering
\includegraphics[width=0.95\textwidth, page=2]{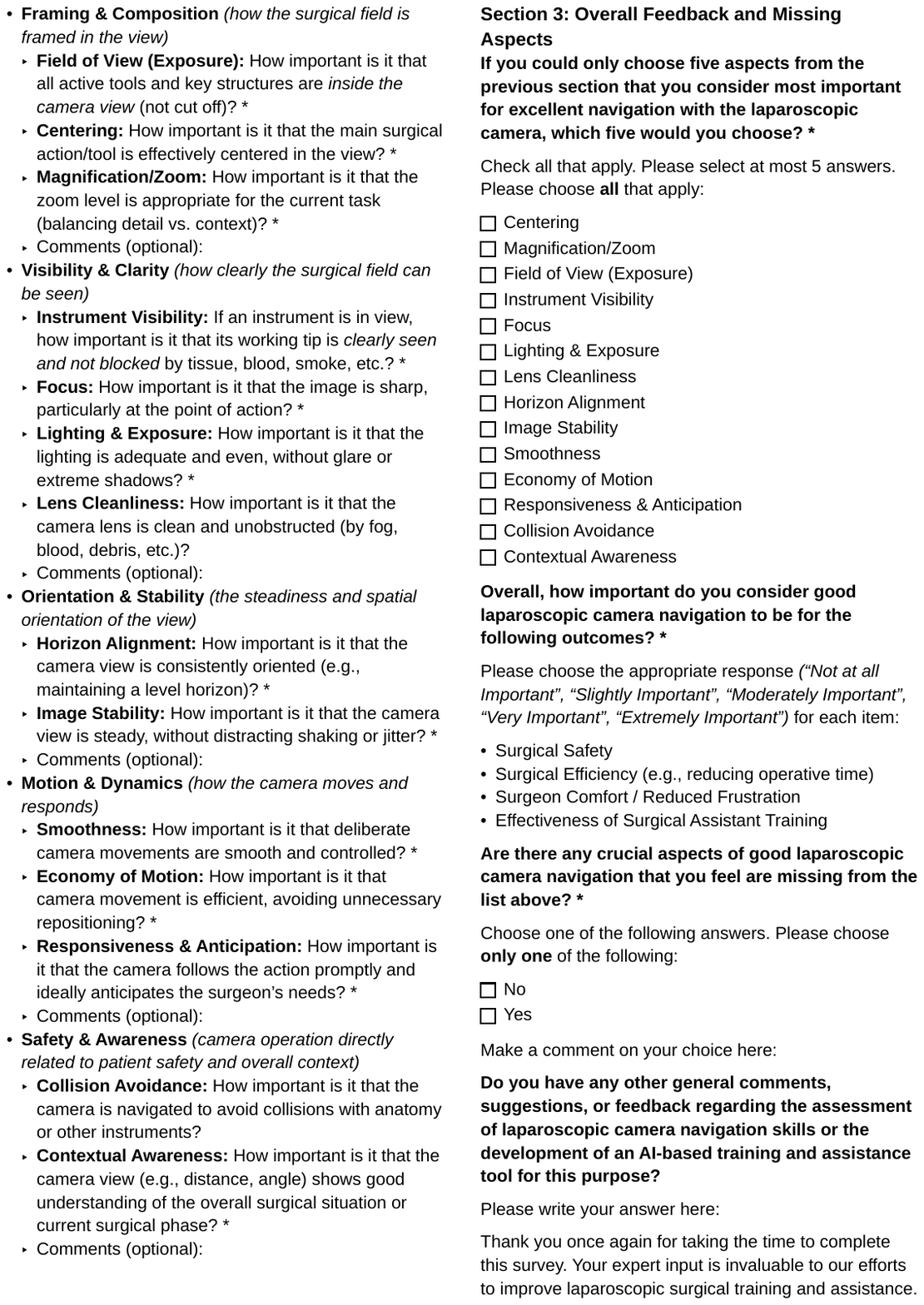}
\caption{Written form of the LimeSurvey online questionnaire, including all original sections and questions (page 2 of 2).}\label{survey.p2}
\end{figure*}
\end{document}